\documentclass[letterpaper,journal]{IEEEtran}
\usepackage{amsmath,amsfonts}
\usepackage{algorithmic}
\usepackage{algorithm}
\usepackage{array}
\usepackage[caption=false,font=normalsize,labelfont=sf,textfont=sf]{subfig}
\usepackage{textcomp}
\usepackage{stfloats}
\usepackage{placeins}
\usepackage{booktabs}
\usepackage{url}
\usepackage{verbatim}
\usepackage{graphicx}
\usepackage{cite}
\usepackage{times}
\usepackage{mathptmx}
\usepackage{newtxtext,newtxmath}
\usepackage[table]{xcolor}   
\usepackage{pgfplots}
\pgfplotsset{compat=1.18}
\definecolor{eatrCyan}{HTML}{70D5DA}
\definecolor{eatrLightBlue}{HTML}{B4D8F0}
\definecolor{eatrTeal}{HTML}{84C4CA}
\begin{document}

\title{EATR-Stereo: Embodiment-Aware Token Routing of Paired Stereo Evidence for Humanoid Vision-Language-Action Control}

\author{\IEEEauthorblockN{Songwei Wu$^{1,2,\ast}$,
Rui Zhao$^{2,\ast}$,
Fan Yang$^{2}$,
Zhongqiang Nie$^{2}$,
Zhiduo Jiang$^{1}$,
Wandong Sun$^{1}$,\\
Yuwei Li$^{2}$,
Jian Hu$^{2}$,
Yang Liu$^{1,\dagger}$,
Hong Liu$^{1}$}\\
\IEEEauthorblockA{$^{1}$Harbin Institute of Technology,
$^{2}$Honor}\\
$^{\ast}$Equal contribution,
$^{\dagger}$Corresponding author}


\maketitle
\thispagestyle{IEEEtitlepagestyle}
\pagestyle{headings}

\begin{abstract}
Long-horizon humanoid vision-language-action (VLA) control with head-mounted stereo cameras requires visual interfaces that can exploit complementary views while maintaining compatibility with pretrained representations. Existing interfaces often discard complementary stereo evidence or fuse additional observations without preserving the native primary-view pathway or adapting auxiliary information to robot embodiment. We present EATR-Stereo, an embodiment-aware token-routing framework that retains primary-view tokens and constructs primary-aligned Cross-View Auxiliary Tokens (CVATs) by allowing primary-view tokens to query the synchronized auxiliary-view token sequence. A body-segmented proprioceptive routing module further conditions token-wise auxiliary usage on robot configuration history, enabling selective incorporation of stereo evidence during action generation. The routed auxiliary stream augments the language and primary-view context of a pretrained VLA while keeping its vision--language model (VLM) frozen. On a 33-DoF physical humanoid with a 37-D proprioceptive state, we evaluate nine configurations in search--approach--grasp--place--return tasks lasting more than 100~s. EATR-Stereo achieves 60.0\% full-task success, 100.0\% grasp success, and 80.0\% stage success. Under severe asymmetric occlusion, EATR-Stereo achieves 80\% recovery, compared with 30\% for the CVAT baseline, which uses CVATs without state routing. Ablation studies further show the importance of the dual-stream construction and body-segmented proprioceptive routing. These results demonstrate that selectively routed paired stereo evidence supports more reliable spatially grounded control in long-horizon humanoid VLA tasks.
\end{abstract}


\section{INTRODUCTION}

Vision-language-action (VLA) models transfer large-scale vision--language representations to language-conditioned control~\cite{rt2,openvla}, and recent systems extend them to continuous actions and whole-body humanoid behavior~\cite{cogact,gr00t}. Bipedal deployment, however, couples perception to whole-body motion: gait and torso disturbances reach the head-mounted cameras, head and waist motion continually alter viewpoint, and moving arms and hands create configuration-dependent self-occlusion~\cite{humanoidvisuomotor,humanoidcamerastabilization}. In a long horizon task, missed evidence at one stage can propagate through the remainder.

Synchronized stereo provides complementary evidence for this regime. Its two cameras observe the same instant from a fixed or calibratable baseline, supporting correspondence and disparity while providing complementary visibility across views, so that content occluded in one view may remain visible in the other~\cite{stereopolicy,peafowl}. This differs from arbitrary multi-view imagery, which need not share timing, baseline, or scene content. The useful signal is therefore the paired relation, not merely a second image.

Current VLA interfaces exploit only part of this signal. A monocular input has neither binocular disparity nor an alternate observation of an occluded region~\cite{stereopolicy}; independent token concatenation retains both images but does not enforce cross-view-consistent spatial reasoning~\cite{peafowl}. StereoPolicy explicitly relates the views, but its VLA variant replaces the original image tokens with a fused stereo representation, thereby discarding the pretrained monocular token structure~\cite{stereopolicy}. SpatialVLA instead predicts depth from each RGB observation with ZoeDepth and back-projects pixels into camera-frame 3D. Its reported physical evaluations use arm manipulators with static or tripod-mounted cameras~\cite{spatialvla}. They therefore do not cover head-mounted bipedal perception, where monocular depth may be temporally inconsistent and gait disturbs the cameras~\cite{kopf2021robust,videodepthanything,humanoidcamerastabilization}. Existing visual--proprioceptive fusion likewise does not apply body-segmented proprioceptive routing to paired-view tokens, despite configuration-dependent viewpoint and occlusion~\cite{hpt,revip,stereopolicy,peafowl}.

\begin{figure*}[t]
    \centering
    \vspace*{3pt}
    \includegraphics[width=0.9\textwidth,trim=20pt 20pt 20pt 20pt,
        clip]{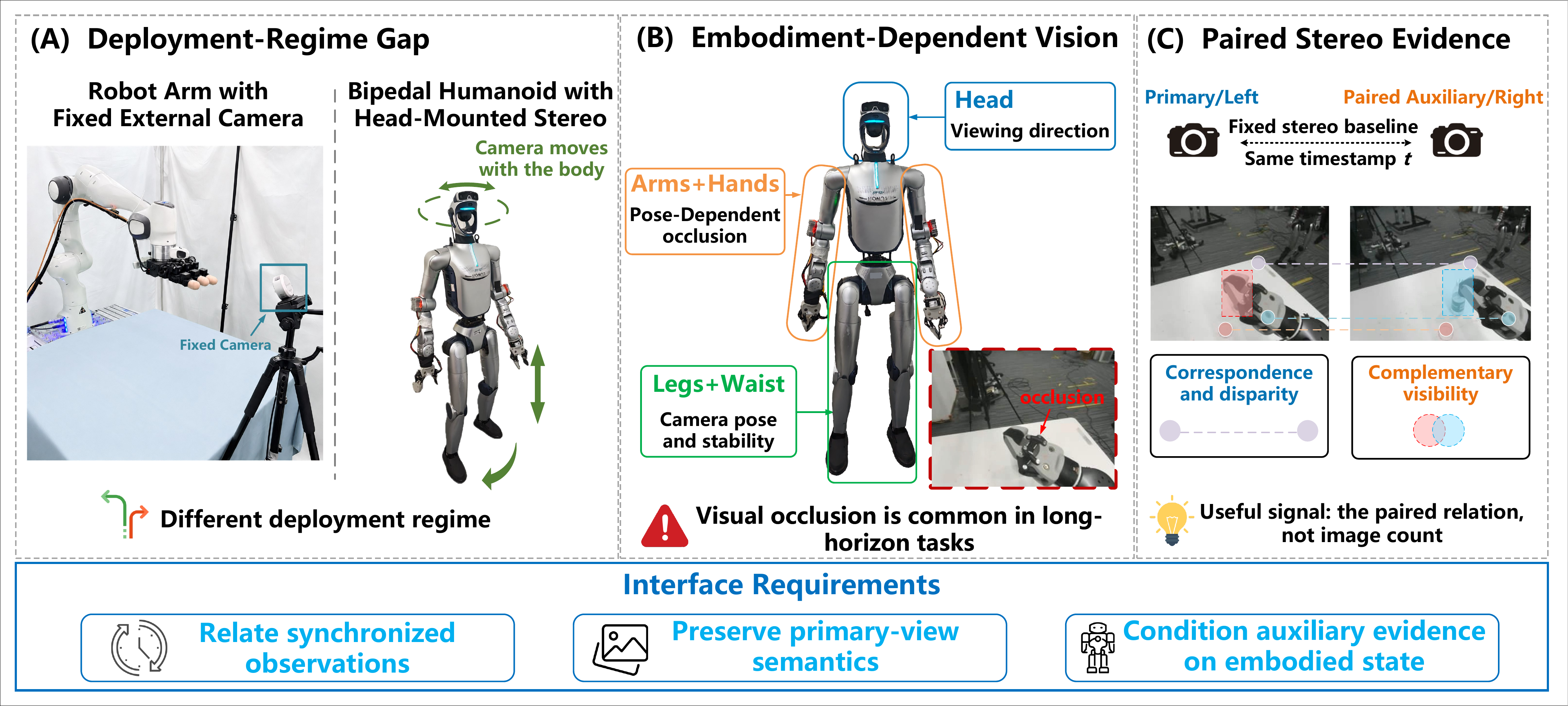}
    \caption{Motivation for paired stereo in humanoid VLA control. (A) Stationary-camera arm settings do not capture the head-camera dynamics of bipedal whole-body motion. (B) Whole-body motion makes head-mounted visual evidence configuration-dependent and consequential over long horizons. (C) Synchronized stereo supplies paired geometric and visibility cues under these dynamics. These conditions motivate an interface that relates synchronized observations, preserves primary-view semantics, and applies body-segmented proprioceptive routing to auxiliary evidence.}
    \label{fig:stereo_design_comparison}
\end{figure*}

We therefore propose \textbf{EATR-Stereo}, an Embodiment-Aware Token Routing framework for pretrained humanoid VLAs. It preserves the primary-view tokens, queries the synchronized auxiliary view to construct primary-aligned Cross-View Auxiliary Tokens (CVATs), and applies body-segmented proprioceptive routing based on recent state history and cross-view relations. The routed auxiliary stream augments the original language and primary-view context before the shared continuous-action expert, while the vision--language model remains frozen. Fig.~\ref{fig:stereo_design_comparison} summarizes the paired sensing setting and the resulting interface requirements.

Our contributions are:
\begin{itemize}
\item A paired-stereo VLA interface that constructs primary-aligned CVATs through primary-query cross-view attention.
\item Body-segmented proprioceptive routing that conditions token-wise CVAT usage on recent state history and cross-view relations.
\item A nine-configuration study on a 33-DoF humanoid in over-100-s tasks. EATR-Stereo reaches 60.0\% full-task and 100.0\% grasp success; under severe occlusion, recovery is 80\% versus 60\% for flat-state routing and 30\% for CVAT alone, with mean successful grasp time reduced from 41.7~s to 22.4~s relative to flat-state routing.
\end{itemize}

\section{RELATED WORK}

\subsection{Vision-Language-Action Models}

Vision-language-action (VLA) models leverage pretrained vision--language representations to improve semantic generalization in robot control. Representative approaches either cast actions into the token space or learn generalist policies across heterogeneous robot data, including RT-2, OpenVLA, and Octo~\cite{rt2,openvla,octo}. More recent VLAs increasingly decouple semantic perception from continuous action generation by conditioning a diffusion- or flow-based action expert on vision--language model (VLM) features~\cite{cogact,pi0}. This paradigm extends to heterogeneous co-training and semantic prediction in $\pi_{0.5}$~\cite{pi05}, and to multi-embodiment and whole-body humanoid control in GR00T N1 and LingBot-VLA 2.0~\cite{gr00t,lingbotvla2}. Despite these advances, existing VLAs largely treat visual observations as conventional semantic inputs and do not explicitly exploit the correspondence and complementary visibility of synchronized stereo views.

\subsection{Stereo and Multi-View Robot Learning}

Robot policies incorporate spatial information through explicit 3D representations, depth-enhanced visual features, and multi-view observations. Geometric approaches operate on point clouds or rendered 3D views~\cite{dp3,cola,rvt}, while depth-based methods augment 2D representations with monocular geometric priors~\cite{depthanything,spatialvla}. Multi-view policies exploit complementary evidence across observations, including geometry-grounded inputs and synthesized auxiliary views~\cite{obeyedvla,multiviewlatent}. Most closely related, StereoPolicy processes synchronized stereo images with shared 2D encoders and cross-view attention, enabling implicit geometric reasoning without explicit disparity or point-cloud reconstruction~\cite{stereopolicy}. By contrast, EATR-Stereo places a primary-aligned auxiliary stream alongside the native image-token pathway, enabling selective use of cross-view evidence.

\subsection{Reliability-Aware and Proprioception-Conditioned Fusion}

Robot policies must integrate visual evidence with proprioceptive state while remaining robust to observation changes. Existing approaches align vision and state within shared representations, adaptively rebalance their contributions, or exploit robot configuration for adaptive control~\cite{hpt,revip,ma2026robotic}, while other methods improve robustness to corrupted observations through filtering and perturbation-based training~\cite{stablevla,strongvla}. EATR-Stereo instead uses body-segmented proprioceptive routing: segment-wise state features and primary--auxiliary visual relations modulate CVATs according to the robot configuration.

In summary, EATR-Stereo targets synchronized stereo, preserves the pretrained primary representation through a separate, primary-aligned CVAT stream, and applies body-segmented proprioceptive routing to selectively incorporate auxiliary evidence.

\section{METHODS}
\label{sec:method}

\subsection{Problem Formulation}
\label{subsec:problem}

We formulate long-horizon humanoid manipulation as the prediction of continuous action chunks conditioned on synchronized stereo observations, language instructions, and proprioceptive history. At time $t$, the policy observation is
\begin{equation}
    o_t=\left(I_t^L,I_t^R,\ell_t,S_t\right),\qquad
    S_t=\left(s_{t-K+1},\ldots,s_t\right),
    \label{eq:observation}
\end{equation}
where $(I_t^L,I_t^R)$ is a synchronized stereo pair, $\ell_t$ is the current language instruction, and $s_t$ denotes the robot proprioceptive state at a single control step. Thus, $S_t$ contains the latest $K$ proprioceptive observations.

The policy predicts an $H$-step continuous-action chunk
\begin{equation}
    \widehat{\mathcal{Y}}_t
    =\left(\hat{\mathbf a}_t,\ldots,\hat{\mathbf a}_{t+H-1}\right)
    =\pi_\theta(I_t^L,I_t^R,\ell_t,S_t),
    \label{eq:policy}
\end{equation}
where $\hat{\mathbf a}_t$ is the predicted continuous robot control command at one execution step. We use $\mathcal{Y}_t=(\mathbf a_t,\ldots,\mathbf a_{t+H-1})$ to denote the corresponding target action chunk.

\begin{figure*}[t]
    \centering
    \vspace*{3pt}
    \includegraphics[width=0.9\textwidth,trim=20pt 20pt 20pt 20pt,
        clip]{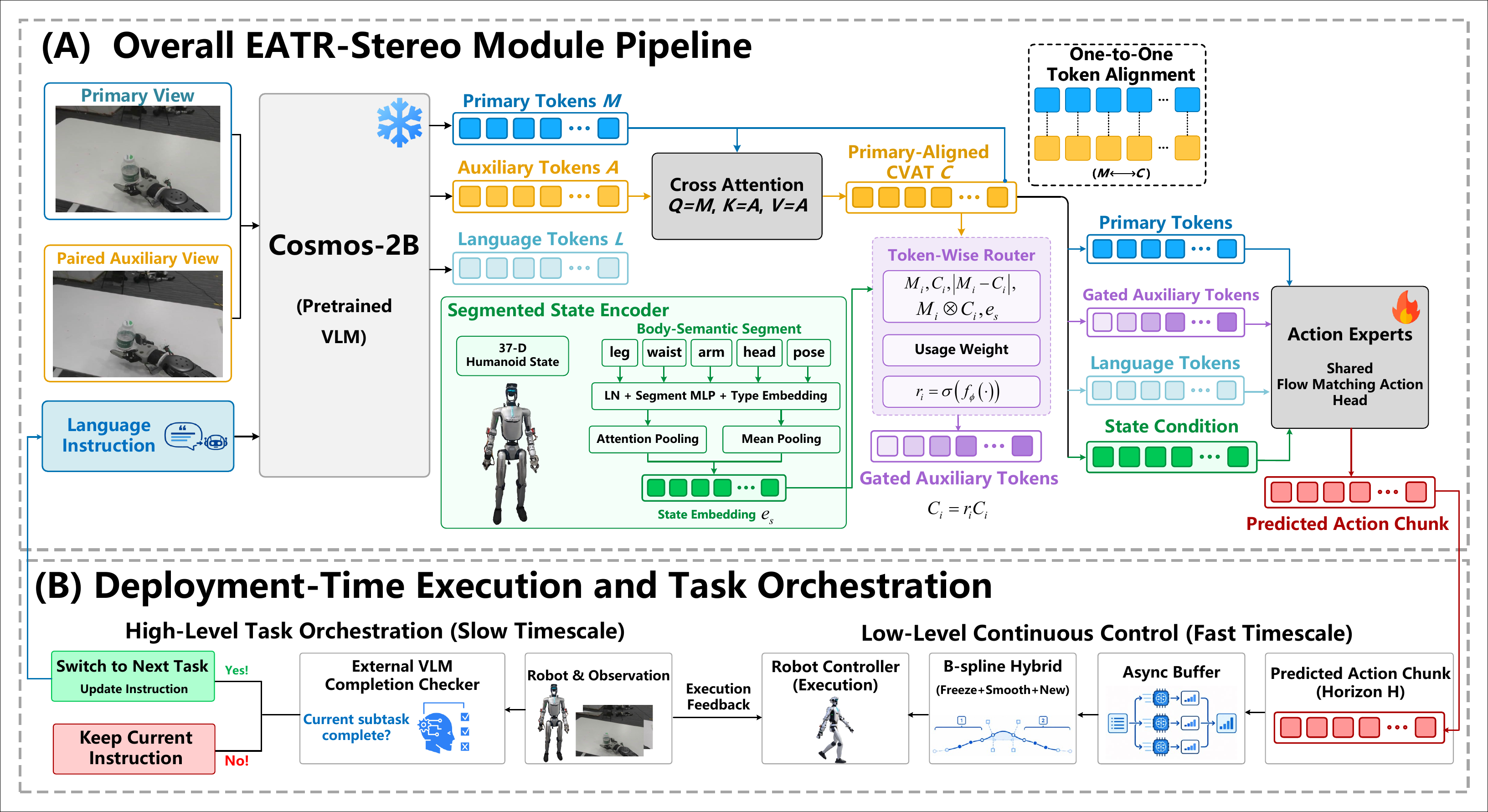}
    \caption{Overview of EATR-Stereo and its deployment pipeline. (A) A frozen pretrained VLM encodes the synchronized primary and auxiliary views with the language instruction. Primary-query cross-view attention constructs a separate, primary-aligned CVAT stream while preserving the native primary tokens. Body-segmented proprioceptive routing conditions token-wise CVAT usage before a shared flow-matching action head predicts an action chunk. (B) At deployment, an external VLM evaluates subtask completion and switches the instruction through a slow-timescale loop, while asynchronous buffering and local B-Spline stitching support fast-timescale control. These deployment components lie outside learned stereo routing.}
    \label{fig:method_overview}
\end{figure*}

From each synchronized pair, we designate one image as the primary view $I_t^p$ and the other as the paired auxiliary view $I_t^a$, and keep this assignment fixed during training and deployment. The primary view provides the reference indexing for the auxiliary CVAT stream, as illustrated in Fig.~\ref{fig:method_overview}(A).

\subsection{Cross-View Auxiliary Token Construction}
\label{subsec:cvat}

As illustrated in Fig.~\ref{fig:method_overview}(A), the Cross-View Auxiliary Token (CVAT) module constructs a primary-indexed auxiliary visual stream. The synchronized stereo pair and language instruction are provided to the frozen Cosmos VLM~\cite{gr00t}, with the instruction represented by the token sequence $L_t$. Let $E_{\mathrm C}$ denote the VLM module responsible for processing image inputs; it produces the primary and auxiliary visual sequences,
\begin{equation}
    M_t=E_{\mathrm C}(I_t^p),\qquad
    A_t=E_{\mathrm C}(I_t^a),
    \label{eq:visual_tokens}
\end{equation}
where $M_t,A_t\in\mathbb{R}^{N\times d}$ share the same VLM hidden space. The primary sequence $M_t$ is retained unchanged along the native visual pathway, while the auxiliary sequence $A_t$ is used to construct the CVAT stream.

To associate auxiliary evidence with each primary token, the CVAT module follows StereoPolicy's multi-head cross-view attention design~\cite{stereopolicy}: the primary sequence supplies queries, and the auxiliary sequence supplies keys and values. For head $h$,
\begin{equation}
    \begin{aligned}
    Q_h &= \operatorname{LN}(M_t)W_h^Q,\\
    K_h &= \operatorname{LN}(A_t)W_h^K,\\
    V_h &= \operatorname{LN}(A_t)W_h^V,
    \end{aligned}
    \label{eq:cvat_qkv}
\end{equation}
and the corresponding head output is
\begin{equation}
    O_h=
    \operatorname{softmax}\left(
    \frac{Q_hK_h^\top}{\sqrt{d_h}}
    \right)V_h,
    \label{eq:cvat_head}
\end{equation}
where $d_h$ is the head dimension.

We concatenate and project the outputs of all $N_h$ heads and then apply a token-wise feed-forward residual block:
\begin{equation}
    C_t^{(0)}
    =
    \operatorname{Concat}_{h=1}^{N_h}(O_h)W^O,
    \quad
    C_t
    =
    C_t^{(0)}
    +
    \operatorname{MLP}\left(
    \operatorname{LN}(C_t^{(0)})
    \right),
    \label{eq:cvat_output}
\end{equation}
yielding $C_t\in\mathbb{R}^{N\times d}$, whose token indexing follows the primary-query sequence $M_t$. The CVAT stream $C_t$ is then passed to the token router.

\subsection{Body-Segmented Proprioceptive Routing}
\label{subsec:routing}

As illustrated in Fig.~\ref{fig:method_overview}(A), the body-segmented proprioceptive router conditions token-wise CVAT usage on a segmented representation of recent proprioceptive history. We partition the proprioceptive state into $G=5$ semantic groups: legs, arms, head, waist, and robot pose, where the arms group includes both arm and hand states. The dimensionality of each group may vary across platforms.

For each segment $g$, its $K$-frame history is flattened, normalized by LayerNorm, and mapped to a common feature dimension by an independent lightweight MLP, followed by a learned segment-type embedding:
\begin{equation}
    z_{t,g}
    =
    f_g\!\left(
    \operatorname{LN}_g\!\left(
    \operatorname{Flatten}(s_{t-K+1:t,g})
    \right)
    \right)
    + e_g,
    \quad g=1,\ldots,G,
    \label{eq:segment_encoding}
\end{equation}
where $z_{t,g}\in\mathbb{R}^{d_s}$ denotes the resulting segment representation.

The segment representations are aggregated through attention-weighted and mean pooling:
\begin{equation}
\begin{aligned}
    \alpha_{t,g}
    &=
    \frac{\exp(f_{\mathrm{attn}}(z_{t,g}))}
    {\sum_{j=1}^{G}\exp(f_{\mathrm{attn}}(z_{t,j}))}, \\
    h_t^s
    &=
    f_{\mathrm{fuse}}\!\left(
    \left[
    \sum_{g=1}^{G}\alpha_{t,g}z_{t,g};
    \frac{1}{G}\sum_{g=1}^{G}z_{t,g}
    \right]
    \right).
\end{aligned}
\label{eq:state_aggregation}
\end{equation}

For each token index $i$, the router jointly uses the primary token $M_{t,i}$, the corresponding CVAT token $C_{t,i}$, their element-wise feature relationships, and the shared proprioceptive condition $h_t^s$. Specifically,
\begin{equation}
\begin{aligned}
    \bar M_{t,i} &= \operatorname{LN}(M_{t,i}),
    \qquad
    \bar C_{t,i} = \operatorname{LN}(C_{t,i}), \\
    q_{t,i}
    &=
    \left[
    \bar M_{t,i};
    \bar C_{t,i};
    |\bar M_{t,i}-\bar C_{t,i}|;
    \bar M_{t,i}\odot\bar C_{t,i};
    h_t^s
    \right],
\end{aligned}
\label{eq:routing_descriptor}
\end{equation}
where $|\cdot|$ and $\odot$ denote element-wise absolute difference and product, respectively.

A lightweight routing MLP predicts one scalar gate for each CVAT token:
\begin{equation}
    r_{t,i}
    =
    \sigma\!\left(f_r(q_{t,i})\right),
    \qquad
    \widetilde C_{t,i}
    =
    r_{t,i}C_{t,i},
    \label{eq:token_routing}
\end{equation}
where $r_{t,i}\in(0,1)$ and $\widetilde C_t\in\mathbb{R}^{N\times d}$. The router scales only the CVAT stream: $M_t$ contributes to the routing descriptor but remains unmodified. The routed auxiliary tokens $\widetilde C_t$ and the primary tokens subsequently meet only in the context supplied to the action expert.

\subsection{VLA Integration, Training, and Deployment}
\label{subsec:vla_integration}

As shown in Fig.~\ref{fig:method_overview}(A), the routed auxiliary tokens are appended to the language-token sequence and the preserved primary-token sequence:
\begin{equation}
    E_t^{\mathrm{VLA}}
    =
    \left[L_t;M_t;\widetilde C_t\right].
    \label{eq:vla_context}
\end{equation}
The resulting context is supplied to the GR00T1.7 action expert together with the original robot-state condition $S_t$; the segmented state representation introduced in Sec.~\ref{subsec:routing} is used only for auxiliary-token routing and does not replace this state pathway. Following the standard GR00T1.7 formulation~\cite{gr00t}, the action expert are trained with the flow-matching objective
\begin{equation}
    \mathcal{L}_{\mathrm{FM}}
    =
    \mathbb{E}_{t,\tau}
    \left[
    \left\|
    v_\theta
    \left(
    \mathcal{Y}_t^\tau,\tau
    \mid E_t^{\mathrm{VLA}},S_t
    \right)
    -
    u_\tau
    \right\|_2^2
    \right],
    \label{eq:flow_loss}
\end{equation}
where $\mathcal{Y}_t^\tau$ is the interpolated state of the target action chunk $\mathcal{Y}_t$ at flow time $\tau$, $u_\tau$ is the target velocity field, and $v_\theta$ is the predicted field. The Cosmos VLM remains frozen, while the CVAT module, segmented state encoder, token router, and GR00T action expert are jointly optimized.

During deployment, action chunks are predicted asynchronously and placed in an execution buffer. At each transition, the hybrid trajectory retains the committed prefix, replaces a local window with a cubic B-Spline, and appends the untouched suffix of the new chunk. The transition spline is
\begin{equation}
    B(\eta)
    =
    \sum\nolimits_{i=0}^{n}
    N_{i,3}(\eta)\mathbf{c}_i,
    \label{eq:bspline}
\end{equation}
where $\eta$ is local time, $N_{i,3}$ is the cubic B-Spline basis function indexed by $i$, and $\{\mathbf c_i\}_{i=0}^{n}$ are the $n+1$ action-space coefficient vectors. The spline matches the action and velocity at the stitching boundary and inherits its terminal value and finite-difference velocity from the new transition window. A short cubic-Hermite adjustment precedes uniform waypoint sampling, including the window endpoint, and independent spline fitting along each action dimension.

At the task level, an external VLM evaluates subtask completion from visual observations and produces a binary completion signal. The current instruction is retained until completion is detected, after which the VLM switches to the next subtask instruction, which is encoded into $L_t$ for subsequent policy inference. This slow-timescale module does not participate in low-level action generation.

\section{EXPERIMENTS}
\label{sec:experiments}

\subsection{Experimental Questions and Setup}
\label{subsec:exp_setup}

We evaluate EATR-Stereo along four dimensions:
(i) long-horizon physical-humanoid performance,
(ii) contributions of its representation and routing components,
(iii) generalization to position shifts and language-conditioned target selection, together with robustness to severe asymmetric occlusion, and
(iv) whether the relative method ranking transfers to larger-scale arm-based simulation.

\begin{figure*}[!t]
    \centering
    \vspace*{3pt}
    \includegraphics[width=0.9\textwidth,trim=20pt 20pt 20pt 20pt,
        clip]{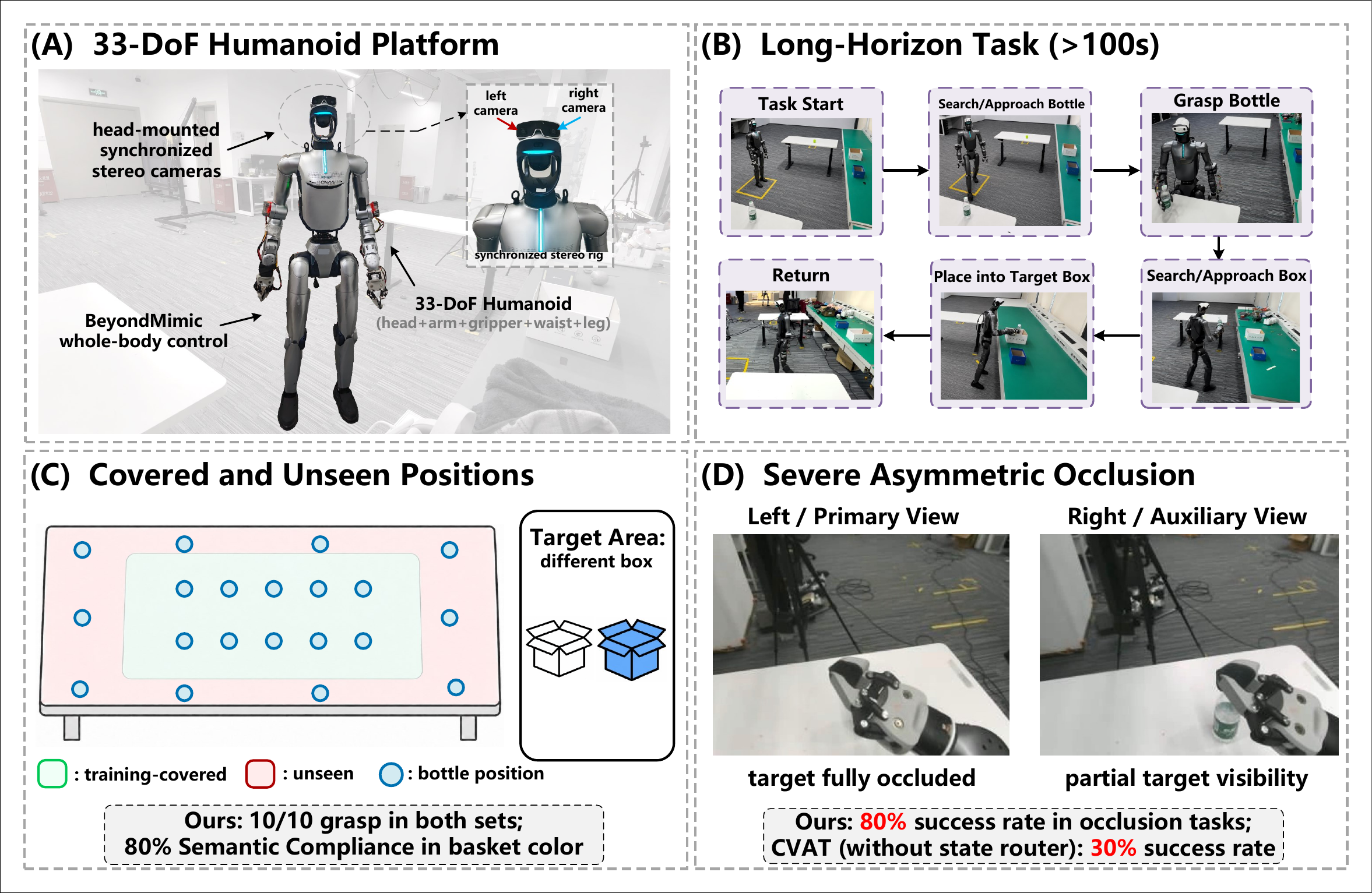}
    \caption{Real-humanoid long-horizon evaluation under a safety-assisted protocol allowing at most two interventions per trial. (A) The HONOR Omega 1.0 platform with synchronized head-mounted stereo cameras and BeyondMimic whole-body control. (B) The search--approach--grasp--place--return task, which lasts more than 100~s. (C) Demonstration-covered (ID) and unseen (OOD) bottle positions tested in a fixed order; annotations for EATR-Stereo report 10/10 grasp success in each split and 80\% basket-color compliance. (D) Targeted severe asymmetric-occlusion analysis, where the bottle is fully hidden in the primary view but remains partially visible in the auxiliary view; EATR-Stereo recovers in 8/10 trials, compared with 3/10 for CVAT without state routing.}
    \label{fig:real_experiments}
\end{figure*}

\textit{Training and comparisons:}
For the physical-humanoid experiments, all nine configurations use the same GR00T1.7 backbone~\cite{gr00t}, 1000 segmented-task demonstrations, and identical optimization settings.
The VLM is frozen; each policy uses a six-frame state history and is trained for 60,000 steps with a global batch size of 512 on 16 NVIDIA H20 GPUs.
The baselines include GR00T1.7-Mono, GR00T-Wide, the default two-image token-concatenation GR00T~\cite{gr00t}, and StereoPolicy~\cite{stereopolicy} adapted to the same backbone.
We additionally evaluate Main-XAttn, CVAT, CVAT-Flat, Stereo-Route, and EATR-Stereo.
Main-XAttn updates the primary stream through cross-attention, whereas CVAT preserves the native primary tokens and appends a primary-aligned auxiliary stream without state routing.
CVAT-Flat and EATR-Stereo differ only in flat versus body-segmented proprioceptive encoding, while Stereo-Route replaces CVAT with the stereo-fusion scheme of StereoPolicy.

\textit{Simulation protocol:}
We conduct 3600 Franka-arm rollouts across 18 RoboCasa365 tasks~\cite{robocasa}, with 20 trials per task and configuration.
QwenPI0.5~\cite{pi05} is included only in simulation as an additional external VLA baseline.
All policies are trained from RoboCasa365 demonstrations under the same protocol.
Wrist-camera observations are excluded, leaving only synchronized left and right agent views to isolate paired-view modeling.
Aggregate task success is the primary simulation metric.

\textit{Physical-humanoid protocol:}
Experiments are conducted on a 33-DoF HONOR Omega 1.0 humanoid platform, controlled by BeyondMimic~\cite{beyondmimic}.
The 37-D proprioceptive state comprises the 33 joint dimensions and a 4-D base-orientation quaternion.
The head-mounted stereo rig provides synchronized $240\times320$ images, with left and right views used as primary and auxiliary inputs without camera parameters.
Policies run at 10~Hz on an NVIDIA RTX 4090 and predict 30-step action chunks.
All methods share the same local cubic B-Spline chunk transition ($k=3$), using eight sampled waypoints and eight blending steps.

Each complete task lasts more than 100~s and consists of bottle search, approach, grasp, language-conditioned placement into the instructed colored basket, and return to the initial position [Fig.~\ref{fig:real_experiments}(B)].
We conduct 180 physical trials, with 20 per configuration, evenly split between demonstration-covered (ID) and unseen (OOD) bottle positions [Fig.~\ref{fig:real_experiments}(C)].
Objects are reset before each trial, and all methods share the same external vision--language monitor for binary subtask-completion detection and instruction switching.

For severe asymmetric occlusion [Fig.~\ref{fig:real_experiments}(D)], CVAT, CVAT-Flat, and EATR-Stereo are each evaluated over 10 trials with a 90-s timeout.
The robot hand fully occludes the bottle in the primary view while leaving it partially visible in the auxiliary view.
Recovery requires a successful lift within the timeout; recovery time is measured from placement at the occluded position to the successful lift and averaged over successful trials.

\textit{Metrics:}
Full-task success requires grasping the bottle, placing it in the instructed basket, and returning to the initial position; grasp success requires a successful lift.
Stage success is the mean of grasp and full-task success, with ID and OOD rates computed separately over their respective 10-trial subsets.
Semantic compliance measures whether the selected basket matches the language instruction: the basket on the instructed side for left/right instructions and the basket with the instructed color for color instructions.
Full-task success is the primary physical-world metric.

\subsection{RoboCasa Arm-Based Simulation Validation}
\label{subsec:robocasa}

\begin{table}[!t]
\caption{Top-seven RoboCasa365 results over all 18 tasks.}
\label{tab:robocasa}
\centering
\small
\setlength{\tabcolsep}{4pt}
\begin{tabular}{lcc}
\toprule
Method & Success (\%) $\uparrow$ & Success/Trials \\
\midrule
\textbf{EATR-Stereo} & \textbf{43.33} & \textbf{156/360} \\
CVAT          & 39.44 & 142/360 \\
StereoPolicy  & 38.06 & 137/360 \\
CVAT-Flat     & 37.50 & 135/360 \\
QwenPI0.5 & 36.11 & 130/360 \\
GR00T1.7 & 35.56 & 128/360 \\
\textsc{GR00T1.7-Mono} & 32.78 & 118/360 \\
\bottomrule
\end{tabular}
\end{table}

Before the primary physical-humanoid evaluation, we conduct a large-scale arm-based simulation study on RoboCasa365 to provide preliminary validation over a broader task set. Table~\ref{tab:robocasa} reports aggregate success across all 18 tasks.

EATR-Stereo achieves the highest aggregate success rate of 43.33\% (156/360), outperforming CVAT, StereoPolicy, CVAT-Flat, QwenPI0.5, GR00T1.7, and \textsc{GR00T1.7-Mono} by 3.89, 5.27, 5.83, 7.22, 7.78, and 10.56 percentage points (pp), respectively.
QwenPI0.5 instantiates the $\pi_0.5$ framework~\cite{pi05} with a Qwen-2B~\cite{qwen3} VLM backbone.
The relatively low absolute success rates should be interpreted in light of the restricted observation setting: unlike the original RoboCasa setup~\cite{robocasa}, wrist-camera observations are excluded, while the paired agent views exhibit substantial occlusion in several tasks.
Despite this more challenging visual setting, EATR-Stereo consistently leads the compared methods, providing large-scale simulation support for the proposed cross-view representation and body-segmented proprioceptive routing before their evaluation on the physical humanoid platform.

\subsection{Real-Humanoid Main Comparison}
\label{subsec:real_main}

The primary evaluation measures long-horizon physical-humanoid manipulation under whole-body motion, dynamic self-occlusion, and unseen target configurations. Table~\ref{tab:real_main} compares EATR-Stereo with three GR00T-based baselines~\cite{gr00t} and StereoPolicy~\cite{stereopolicy} in terms of full-task, grasp, and stage success, with stage success further separated into demonstration-covered (ID) and unseen (OOD) positions.

\begin{table*}[!t]
\caption{Main physical-humanoid comparison over 20 trials per method. ID and OOD denote demonstration-covered and unseen bottle positions, respectively.}
\label{tab:real_main}
\centering
\small
\begin{tabular*}{\textwidth}{@{\extracolsep{\fill}}lccccc@{}}
\toprule
Method & Full-task Success $\uparrow$ & Grasp Success $\uparrow$ & Stage Success Rate $\uparrow$ & \shortstack{ID Stage Success $\uparrow$} & \shortstack{OOD Stage Success $\uparrow$} \\
\midrule
\textsc{GR00T1.7-Mono} & 35.0\% & 55.0\% & 45.0\% & 55.0\% & 35.0\% \\
\textsc{GR00T-Wide}    & 20.0\% & 45.0\% & 32.5\% & 40.0\% & 25.0\% \\
\textsc{GR00T}         & 35.0\% & 80.0\% & 57.5\% & 75.0\% & 40.0\% \\
StereoPolicy            & 45.0\% & 85.0\% & 65.0\% & 80.0\% & 50.0\% \\
\textbf{EATR-Stereo}   & \textbf{60.0\%} & \textbf{100.0\%} & \textbf{80.0\%} & \textbf{90.0\%} & \textbf{70.0\%} \\
\bottomrule
\end{tabular*}
\end{table*}

EATR-Stereo leads every metric in Table~\ref{tab:real_main}. It completes 12 of 20 trials end-to-end (60.0\%), outperforming StereoPolicy by 15.0~pp, \textsc{GR00T1.7-Mono} and \textsc{GR00T} by 25.0~pp, and \textsc{GR00T-Wide} by 40.0~pp. It also grasps the bottle in all 20 trials, a 15.0--55.0~pp improvement over the baselines. The lower success of \textsc{GR00T1.7-Mono} mainly arises during approach, where the robot more frequently deviates from the bottle-directed walking trajectory. \textsc{GR00T-Wide} performs worst overall, which likely reflects the frozen VLM's limited pretraining exposure to wide-aspect-ratio images.

EATR-Stereo's overall stage success is 80.0\%, versus 65.0\% for StereoPolicy and 32.5--57.5\% for the GR00T-based baselines. EATR-Stereo achieves 90.0\% and 70.0\% stage success on ID and OOD positions, respectively; its 20.0~pp OOD advantage over StereoPolicy indicates stronger generalization to unseen bottle positions.

\subsection{Real-Humanoid Ablations and Targeted Analyses}
\label{subsec:real_ablation}

\begin{figure}[!t]
\centering
\includegraphics[width=\columnwidth]{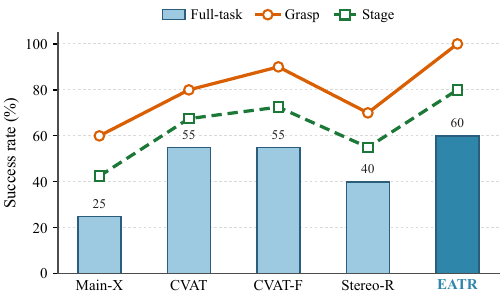}
\caption{Physical-humanoid ablations over 20 trials per method. CVAT-F denotes CVAT-Flat and Stereo-R denotes Stereo-Route.}
\label{fig:real_ablation}
\end{figure}

\textit{Component ablations:}
We ablate three design choices: retaining the native primary-view stream, selectively routing auxiliary-view evidence, and using body-segmented proprioceptive encoding. Fig.~\ref{fig:real_ablation} reports full-task success as bars and grasp and stage-success rates as curves.

Replacing direct modification of the primary tokens (Main-XAttn) with the CVAT dual-stream representation raises full-task success from 25.0\% to 55.0\%, a 30.0~pp gain (Fig.~\ref{fig:real_ablation}). This result supports retaining the pretrained primary representation while introducing auxiliary evidence through a dedicated stream.

CVAT and CVAT-Flat both achieve 55.0\% full-task success, but flat state conditioning improves grasp and stage success from 80.0\% to 90.0\% and from 67.5\% to 72.5\%, respectively. Body-segmented proprioceptive routing further yields full-task, grasp, and stage success rates of 60.0\%, 100.0\%, and 80.0\%, respectively. Stereo-Route achieves 40.0\% full-task and 55.0\% stage success, while EATR-Stereo further improves these metrics through the joint use of preserved primary tokens, the CVAT auxiliary stream, and body-segmented proprioceptive routing. Overall, the ablations show complementary gains from representation preservation, auxiliary-view integration, and body-segmented proprioceptive routing.

\textit{Position and semantic behavior:}

Table~\ref{tab:position_semantic} evaluates position-wise grasp generalization and semantic compliance. EATR-Stereo is the only method with 100\% grasp success on both ID and OOD positions. Its OOD grasp rate exceeds CVAT-Flat by 20.0~pp, StereoPolicy by 30.0~pp, and CVAT and \textsc{GR00T} by 40.0~pp. The resulting zero ID--OOD gap indicates that grasp performance is preserved at bottle positions absent from the demonstrations. At unseen positions, EATR-Stereo also maintains accurate bottle-directed locomotion during approach, reducing positioning errors that are a major source of subsequent grasp failures and suggesting that paired-view integration provides stronger spatial grounding for humanoid navigation and manipulation.

\begin{table*}[!t]
\caption{Position generalization and semantic compliance on the physical humanoid. Grasp success is measured over 10 trials per position split. Semantic compliance is measured over 10 trials per instruction type and denotes selection of the basket on the instructed side (Left/Right) or with the instructed color (Color). Gap denotes the ID-to-OOD drop in grasp success.}
\label{tab:position_semantic}
\centering
\small
\begin{tabular*}{\textwidth}{@{\extracolsep{\fill}}lccccc@{}}
\toprule
& \multicolumn{3}{c}{Position Generalization} & \multicolumn{2}{c}{Semantic Compliance} \\
\cmidrule(lr){2-4}\cmidrule(lr){5-6}
Method & ID Grasp (\%) $\uparrow$ & OOD Grasp (\%) $\uparrow$ & Gap (pp) $\downarrow$ & Left/Right (\%) $\uparrow$ & Color (\%) $\uparrow$ \\
\midrule
\textsc{GR00T1.7-Mono} & 70  & 40  & $30$ & 100 & 80 \\
\textsc{GR00T-Wide}    & 50  & 40  & $10$ & 0   & 0  \\
\textsc{GR00T}         & 100 & 60  & $40$ & 100 & 80 \\
StereoPolicy           & 100 & 70  & $30$ & 60  & 0  \\
Main-XAttn              & 60  & 60  & $0$   & 0   & 0  \\
CVAT                    & 100 & 60  & $40$ & 100 & 80 \\
CVAT-Flat               & 100 & 80  & $20$ & 100 & 80 \\
Stereo-Route            & 80  & 60  & $20$ & 100 & 80 \\
\textbf{EATR-Stereo}   & \textbf{100} & \textbf{100} & \textbf{$0$} & \textbf{100} & \textbf{80} \\
\bottomrule
\end{tabular*}
\end{table*}

For EATR-Stereo, Table~\ref{tab:position_semantic} reports 100\% compliance with left/right instructions and 80\% compliance with color instructions. The method therefore matches the best semantic-compliance performance among the compared methods while providing substantially higher OOD grasp success. The stereo integration improves spatial generalization without degrading language-conditioned behavior.

\textit{Occlusion recovery:}
We next report results for the severe asymmetric-occlusion protocol described above and illustrated in Fig.~\ref{fig:real_experiments}(D).

\begin{table}[!t]
\caption{Severe asymmetric-occlusion recovery. Mean time uses successful trials only; all trials use a 90-s timeout.}
\label{tab:occlusion}
\centering
\small
\setlength{\tabcolsep}{5pt}
\begin{tabular}{lcc}
\toprule
Method & Recovery Success $\uparrow$ & Mean Time (s) $\downarrow$ \\
\midrule
CVAT & 30\% (3/10) & $>55.0$ \\
CVAT-Flat & 60\% (6/10) & 41.7 \\
\textbf{EATR-Stereo} & \textbf{80\% (8/10)} & \textbf{22.4} \\
\bottomrule
\end{tabular}
\end{table}

As reported in Table~\ref{tab:occlusion}, EATR-Stereo recovers in 80\% of trials, compared with 60\% for CVAT-Flat and 30\% for CVAT. Its 22.4-s mean recovery time is 46.3\% lower than that of CVAT-Flat and at least 59.3\% lower than that of CVAT. These results demonstrate faster and more reliable closed-loop recovery with body-segmented proprioceptive routing.

\textit{System-level training cost:}
Finally, we compare training cost and physical-task performance against the default two-image \textsc{GR00T} baseline and an Estimated-Depth VLA based on Depth Anything V3~\cite{depthanything}. All methods use the same dataset, base VLA, 60,000-step schedule, global batch size of 512, and 16 NVIDIA H20 GPUs. \textsc{GR00T} and EATR-Stereo keep the VLM frozen. Following Spatial Forcing~\cite{Spatial-Forcing}, the estimated-depth method fine-tunes four VLM layers to align VLM visual tokens from the two RGB streams with tokens produced by Depth Anything V3.

\begin{table}[!t]
\caption{System-level training cost and physical-task performance.}
\label{tab:training_cost}
\centering
\footnotesize
\setlength{\tabcolsep}{3pt}
\begin{tabular}{lcccc}
\toprule
Method & VLM Adapt. & Time (h) & Relative & Stage (\%) $\uparrow$ \\
\midrule
\textsc{GR00T} & Frozen & 10.35 & $1.00\times$ & 57.5 \\
\textbf{EATR-Stereo} & Frozen & \textbf{10.95} & \textbf{$1.06\times$} & \textbf{80.0} \\
Estimated-Depth VLA & 4 layers & 41.80 & $4.04\times$ & \textbf{80.0} \\
\bottomrule
\end{tabular}
\end{table}

Table~\ref{tab:training_cost} shows that EATR-Stereo reaches an 80.0\% stage success rate while requiring 10.95~h of training, only 0.60~h (5.8\%) more than \textsc{GR00T}, whose stage success is 57.5\%. The Estimated-Depth VLA also reaches 80.0\% stage success but requires 41.8~h, corresponding to $4.04\times$ the training cost of \textsc{GR00T} and $3.82\times$ that of EATR-Stereo. Thus, EATR-Stereo attains the same stage-level performance as explicit estimated-depth adaptation with substantially lower training overhead.

\textit{Action-chunk continuity:}
\begin{figure}[!t]
\centering
\includegraphics[width=\columnwidth]{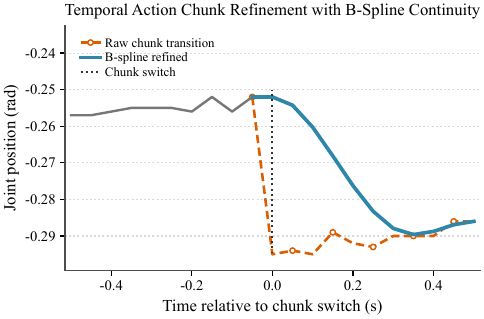}
\caption{Qualitative action-chunk continuity for a representative joint around a chunk boundary.}
\label{fig:bspline_continuity}
\end{figure}

Fig.~\ref{fig:bspline_continuity} shows a representative joint trajectory around an action-chunk boundary. Directly switching to the raw prediction produces an abrupt position change at the boundary, whereas local cubic B-Spline refinement continues from the committed trajectory and gradually joins the new chunk over the blending window.

\section{CONCLUSION}

This work investigated how synchronized stereo observations can be integrated into a pretrained humanoid VLA while preserving its native primary-view representation. EATR-Stereo introduces a primary-aligned Cross-View Auxiliary Token pathway and body-segmented proprioceptive routing to selectively incorporate complementary stereo evidence. On a 33-DoF humanoid with a 37-D proprioceptive state, EATR-Stereo achieves 60.0\% full-task success, 100.0\% grasp success, and 80.0\% stage success in tasks lasting more than 100~s, demonstrating improved long-horizon manipulation reliability. The proposed method also shows more reliable spatially grounded behavior during locomotion and improves recovery under severe asymmetric occlusion. Ablation studies further highlight the contributions of primary-token preservation, explicit cross-view auxiliary modeling, and body-segmented proprioception for adaptive evidence selection.

These results show that stereo observations can be effectively incorporated into existing VLA systems through lightweight representation interfaces without redesigning the pretrained vision--language backbone. Through body-segmented proprioceptive routing, EATR-Stereo selectively uses complementary visual evidence, while calibrated uncertainty modeling and explicit depth representations remain promising directions for further study. The current evaluation focuses on the Omega humanoid platform and complementary arm-based simulation benchmarks; future work will extend evaluation to larger-scale trials, diverse embodiments, and challenging stereo conditions. Overall, EATR-Stereo provides a practical framework for improving embodied policies under viewpoint variation and partial observability.

\bibliographystyle{IEEEtran}
\bibliography{references}

\vfill
\end{document}